\documentclass[a4paper,twocolumn]{article}

\usepackage{jsaiac}

\usepackage{color}
\usepackage{natbib}
\usepackage{hyperref}
\usepackage{url}
\usepackage{algorithm}
\usepackage{algpseudocode}
\usepackage{graphicx}
\usepackage{amsmath}
\usepackage{booktabs}
\usepackage{float}

\title{Online Learning of Counter Categories and Ratings in PvP Games}

\address{Chiu-Chou Lin, Department of Computer Science, National Yang Ming Chiao Tung University, Hsinchu 30010, Taiwan, dsobscure@outlook.com}

\author{%
Chiu-Chou Lin\first
\and
I-Chen Wu \first \second
}

\affiliate{
\first{} Department of Computer Science, National Yang Ming Chiao Tung University
\and
\second{} Research Center for Information Technology Innovation, Academia Sinica
}

\begin{abstract}
In competitive games, strength ratings like Elo are widely used to quantify player skill and support matchmaking by accounting for skill disparities better than simple win rate statistics. However, scalar ratings cannot handle complex intransitive relationships, such as counter strategies seen in Rock-Paper-Scissors. To address this, recent work introduced Neural Rating Table and Neural Counter Table, which combine scalar ratings with discrete counter categories to model intransitivity. While effective, these methods rely on neural network training and cannot perform real-time updates. In this paper, we propose an online update algorithm that extends Elo principles to incorporate real-time learning of counter categories. Our method dynamically adjusts both ratings and counter relationships after each match, preserving the explainability of scalar ratings while addressing intransitivity. Experiments on zero-sum competitive games demonstrate its practicality, particularly in scenarios without complex team compositions.
\end{abstract}

\def\BibTeX{{\rm B\kern-.05em{\sc i\kern-.025em b}\kern-.08em%
 T\kern-.1667em\lower.7ex\hbox{E}\kern-.125emX}}
\def\JBibTeX{\leavevmode\lower .6ex\hbox{J}\kern-0.15em\BibTeX}
\def\LaTeXe{\LaTeX\kern.15em2$_{\textstyle\varepsilon}$}
\algnewcommand{\LeftComment}[1]{\Statex \(\triangleright\) #1}

\begin{document}
\maketitle


\section{Introduction}

An effective strength measurement is a cornerstone of both games and multi-agent systems, providing quantitative skill information to players and supporting matchmaking mechanisms that influence engagement, fairness, and overall user experience \citep{art_of_game_design}. Beyond games, these mechanisms also play a vital role in training artificial agents, where balanced matches promote robust learning and strategy development \citep{alpha_star}.

To quantify player strength, widely-used rating systems such as Elo, Glicko, Whole-History Rating (WHR), TrueSkill, and Matchmaking Rating (MMR) offer scalar evaluations of individual skill levels \citep{elo, glicko,whr,true_skill,mmr,elo_mmr}. These systems have demonstrated their effectiveness in competitive games, from traditional board games to modern online multiplayer titles. Scalar ratings outperform simple win-rate statistics by accounting for disparities in player skill, thereby creating more balanced and meaningful matches. However, they are fundamentally limited in handling complex win rate intransitivity, such as counter relationships found in games like \textit{Rock-Paper-Scissors} or \textit{Pok\'emon}
, where strategies cyclically dominate one another \citep{game_balance_analysis, nash_entropy_balancing}.

\begin{figure}[t]
    \centering
    \includegraphics[width=0.4\textwidth]{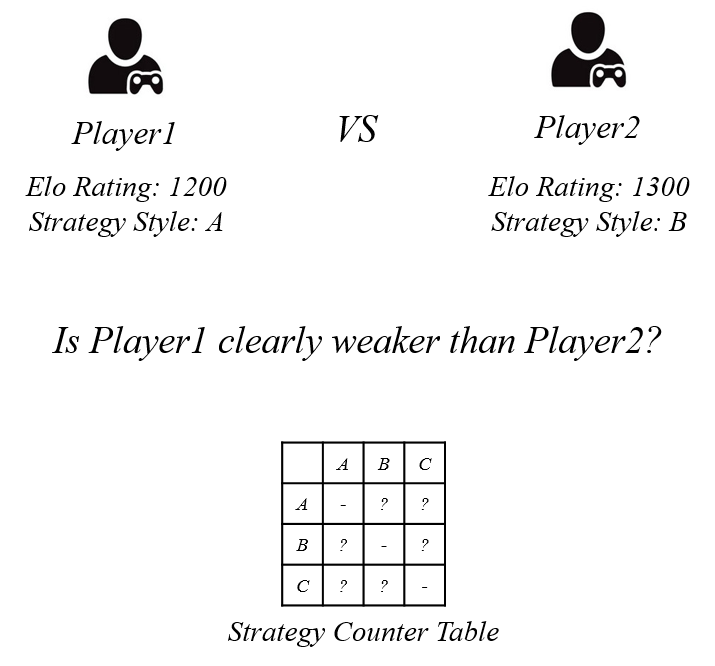}
    \caption{An illustration of counter relationships in games. While Player 2's Elo rating may be slightly higher than Player 1's, the dominance relationship between strategies A and B remains unclear due to potential counter dynamics, emphasizing the limitations of scalar ratings in capturing such complexities.}
    \label{fig:counter_relationships}
\end{figure}

As illustrated in Figure~\ref{fig:counter_relationships}, scalar rating systems often fail to capture intransitive strategy dynamics, which are common in games featuring cyclic dominance among strategies. Such dynamics enrich the strategic diversity of gameplay by encouraging players to explore different approaches. Players naturally favor specific strategies that align with their preferences or expertise, further contributing to the variety of interactions \citep{playstyle_distance,playstyle_similarity}. However, this diversity presents a significant challenge for scalar rating systems, which are built on the assumption of transitive relationships. For instance, in card games like \textit{Hearthstone}, the presence of diverse deck types leads to matchups with non-transitive outcomes, making it difficult for scalar ratings to fully capture the complexity of these interactions.

To address these limitations, researchers have explored multi-dimensional rating systems, which can model intransitivity by representing skill as a vector rather than a scalar \citep{m_elo,nd_ability,hyperbolic_elo}. While these systems are theoretically powerful, they suffer from poor explainability, making it challenging for players to interpret their ratings. Additionally, their complexity hinders their utility for quick balance analysis, where scalar ratings allow straightforward identification of overly strong or weak elements \citep{game_balance_analysis}.

Recent advancements, such as the Neural Rating Table (NRT) and Neural Counter Table (NCT) proposed by \citet{game_balance_analysis}, effectively balance simplicity and precision. Built on the Bradley-Terry model \citep{bradley_terry}, these methods address intransitive win relationships by clustering residuals from scalar rating predictions into a smaller, enumerable counter table. This design introduces counter categories, significantly reducing the complexity of modeling pairwise interactions. For instance, while constructing a pairwise $N \times N$ win rate table requires extensive game records for $N$ players, many strategy games involve far fewer unique strategic genres ($M$). By leveraging this reduction, the $M \times M$ counter table approximates pairwise prediction accuracy while maintaining interpretability. Its effectiveness has been validated through applications in popular games such as \textit{Age of Empires II} and \textit{Hearthstone}.

Despite these strengths, NRT and NCT, along with other neural network-based rating systems \citep{neural_bt_model}, rely on static datasets and are not designed for online updates. Unlike Elo and MMR, which support real-time incremental updates after each match, these methods lack adaptability, limiting their applicability in dynamic environments.

In this paper, we introduce Elo Residual Counter Category learning (Elo-RCC), a novel online update algorithm that extends the traditional Elo rating system to incorporate counter categories. Elo-RCC combines the simplicity and interpretability of scalar ratings with the ability to address win rate intransitivity. By leveraging an expectation-maximization (EM) algorithm \citep{em_algorithm}, our method dynamically learns counter category probabilities for players or strategies based on their best responses to the evolving counter table.

We validate Elo-RCC using the publicly available dataset from \citet{game_balance_analysis}, focusing on games with clear counter relationships and limited strategy complexity, where the use of neural networks for generalization is unnecessary. Our results demonstrate that Elo-RCC achieves comparable performance to NCT while enabling real-time updates, making it a practical alternative for both game balance analysis and matchmaking systems. Furthermore, among online update methods, Elo-RCC consistently delivers the best performance, highlighting its effectiveness in addressing counter relationships dynamically.

\section{Background}

To propose a new rating system, it is essential to first revisit the foundations of existing rating systems, particularly the Bradley-Terry model and the widely used Elo rating system. These models form the basis for quantifying player strength and provide a framework for understanding how strengths translate into win probabilities. Building upon these, we introduce the Neural Rating Table (NRT) and Neural Counter Table (NCT), which serve as the foundation of our proposed system.

\subsection{Bradley-Terry Model and Elo Rating}

A common approach to model a player's strength is to describe it as a single numeric value. However, an important question is how this number relates to the player's actual performance in a competitive match. \citet{bradley_terry} addressed this by proposing a simple yet powerful probabilistic model for two-player zero-sum games. The win probability of player $i$ against player $j$ is given by:
\begin{equation}
\label{bt_win_value}
P(i > j) = \frac{\gamma_i}{\gamma_i + \gamma_j},
\end{equation}
where $\gamma_x$ denotes the positive real-valued strength of player $x$. This formulation intuitively represents how relative strengths affect win probabilities.

To ensure numerical stability and ease of computation, $\gamma_x$ is often reparameterized using an exponential function, leading to:
\begin{equation}
\label{bt_win_value_exp}
P(i > j) = \frac{e^{\lambda_i}}{e^{\lambda_i} + e^{\lambda_j}}.
\end{equation}

With further algebraic transformations, this can be expressed in a form that directly aligns with the commonly used Elo rating system:
\begin{equation}
P(i > j) = \frac{1}{1 + 10^{(R_j - R_i)/400}},
\end{equation}
where $R_i$ and $R_j$ represent the Elo ratings of players $i$ and $j$, respectively \citep{elo}. 

While the Bradley-Terry model provides a static representation of strength, the Elo rating system extends this by introducing an incremental update mechanism. After each game, player ratings are updated using a simple formula:
\begin{equation}
R'_A = R_A + K \cdot (S_A - E_A),
\end{equation}
where $R'_A$ is the updated rating, $S_A$ is the actual result (1 for a win, 0 for a loss), $E_A$ is the expected probability of winning, and $K$ is a constant controlling the update magnitude.

Elo's incremental update mechanism makes it well-suited for real-time applications in competitive games. Its simplicity, transparency, and ease of interpretation have contributed to its widespread adoption. Our proposed rating system builds upon these principles to maintain player-friendliness while incorporating enhanced modeling capabilities.

\subsection{Neural Rating/Counter Table}

Building on the Bradley-Terry model, \citet{game_balance_analysis} proposed the Neural Rating Table (NRT) and Neural Counter Table (NCT) to reduce the complexity of game balance analysis and address the intransitivity problem that the Bradley-Terry model cannot handle. For example, scenarios such as $A > B$, $B > C$, and $C > A$ illustrate the intransitivity of strength relationships, which scalar ratings struggle to model. The NRT and NCT introduce neural networks to capture these relationships and provide a structured approach for counter category learning.

The \textbf{Neural Rating Table (NRT)} is a rating predictor that extends the Bradley-Terry model. While the original model is limited to two-player settings, NRT generalizes to team-based matches by aggregating the players in a team into a single entity. A Siamese neural network is employed to approximate the strength of unseen team combinations based on observed data, significantly reducing the need for exhaustive lookup tables while capturing synergy effects among team members.

\begin{figure}[t]
    \centering
    \includegraphics[width=0.3\textwidth]{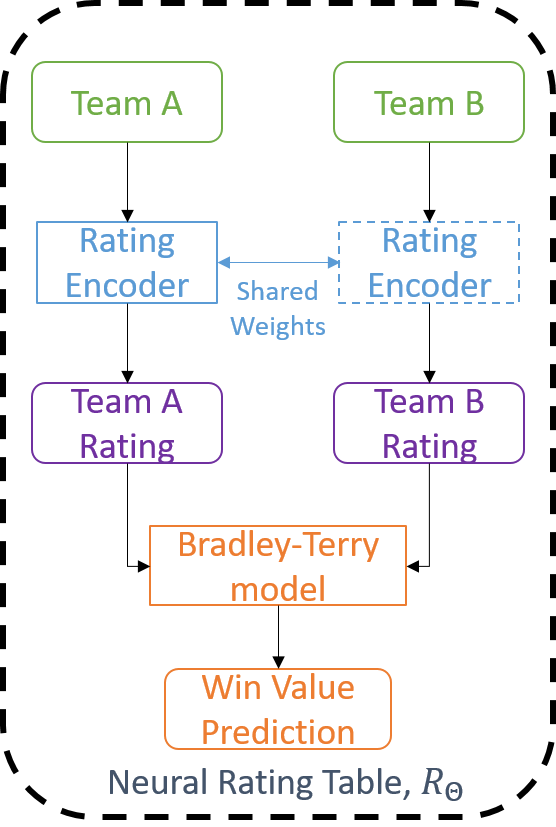}
    \caption{Neural Rating Table (NRT) predicts win probabilities between two teams using shared-weight encoders and the Bradley-Terry model, with exponential activation applied in the rating encoder for reparameterization.}
    \label{figure:neural_rating_table}
\end{figure}

Figure~\ref{figure:neural_rating_table} illustrates the workflow of NRT. The Siamese network, with shared weights \citep{siamese_network}, processes features of two competing teams to predict match outcomes. These features can range from binary agent indices to more sophisticated representations encoding team-level synergies.

Building upon NRT, the \textbf{Neural Counter Table (NCT)} addresses the intransitivity of strength relationships that scalar ratings cannot model. Unlike multi-dimensional rating systems \citep{nd_ability, m_elo, hyperbolic_elo}, NCT simplifies the representation of counter relationships by clustering residual win values $W_{res}$ into discrete counter categories. The residual win value quantifies the deviation between the actual match outcome $W$ and the Bradley-Terry model's prediction, defined as:

\begin{equation} \label{equation:residual_win_value} W_{res}(A, B | R_{\theta}) = W - \frac{R_{\theta}(A)}{R_{\theta}(A) + R_{\theta}(B)}, \end{equation}

where $R_{\theta}(A)$ and $R_{\theta}(B)$ denote the predicted strengths of teams $A$ and $B$, respectively.

\begin{figure}[t] \centering \includegraphics[width=0.3\textwidth]{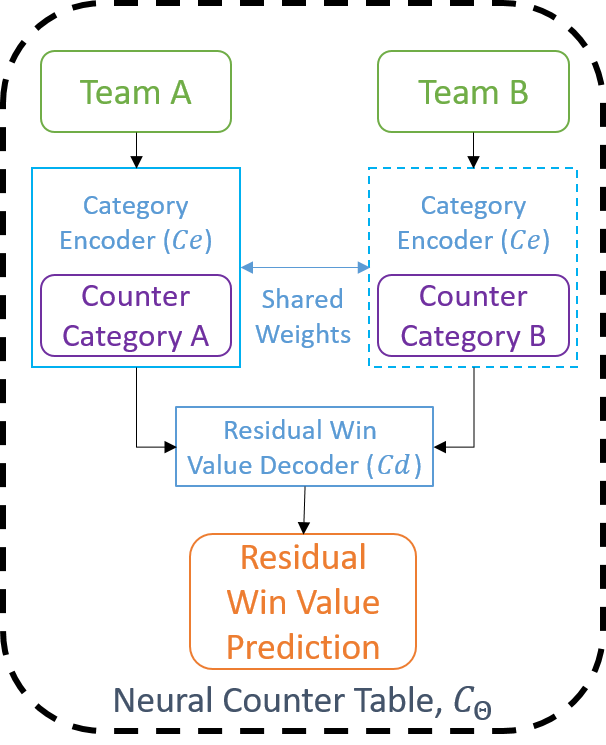} \caption{Neural Counter Table (NCT) models counter relationships using category encoders and residual win value decoders.} \label{figure:neural_counter_table} \end{figure}

Figure~\ref{figure:neural_counter_table} illustrates the workflow of NCT. A neural network maps team compositions into counter categories, leveraging a static dataset and a pretrained NRT model. By employing vector quantization techniques \citep{vq_vae}, NCT clusters residual win values into a finite set of interpretable categories. This process effectively translates complex intransitive relationships into discrete counter categories, offering a domain-agnostic framework for representing strategic interactions.

The final win probability prediction combines both the scalar rating from NRT and the residual adjustment from NCT, expressed as:
\begin{equation} P(A > B) = \frac{R_{\theta}(A)}{R_{\theta}(A) + R_{\theta}(B)} + W_{res}(A, B | C_\theta), \end{equation}
where $W_{res}(A, B | C_\theta)$ is retrieved from the learned counter table $C_\theta$ based on the predicted counter categories of $A$ and $B$. This formulation allows the model to capture intransitive relationships while preserving the interpretability and efficiency of scalar ratings.

While NRT and NCT demonstrate promising results in modeling synergy and counter relationships, they rely on neural networks and static datasets, limiting their suitability for incremental rating systems like Elo. To address this limitation, we propose an online learning algorithm that integrates counter category learning into the Elo framework, allowing for real-time updates and maintaining the simplicity of scalar ratings.

\section{Elo Residual Counter Category}

To design an online update algorithm for NRT and NCT, we clarify that this work does not address the complexity of general team compositions. Although NRT is capable of handling such scenarios to some extent, we simplify the model by focusing solely on individual ratings. Specifically, we adopt the standard Elo rating system to serve as the foundation for our approach, avoiding additional formulations required for team-based settings. While methods like the generalized Bradley-Terry model \citep{generalized_bt} could provide solutions for more complex scenarios, our aim is to propose a straightforward method applicable to individual player settings.

The primary challenge, therefore, lies in implementing an online version of NCT. In particular, we must determine how to assign and update counter categories dynamically for each individual. Below, we introduce our approach and detail the proposed rating system algorithm.

\subsection{Key Ideas}

In the original NCT paper \citep{game_balance_analysis}, the authors first trained an NRT model to establish a stable rating foundation before clustering the residual win values ($W_{res}$). However, in an online setting, obtaining a stable $W_{res}$ prediction is challenging due to the absence of a pre-trained model for reference and the lack of batch outcomes to calculate the average residual values.

To address this limitation, we model a categorical distribution for each individual, initialized uniformly. After each match, player ratings are updated using the Elo formula. Simultaneously, counter categories for both individuals are sampled from their respective distributions. Although these sampled categories may not always be accurate, the sampling mechanism enables exploration to determine the best-fitting category for each individual.

Using the sampled categories and the residual win value derived from the match outcome and Elo ratings, we update the $M \times M$ counter table through tabular regression. Additionally, we maintain an expected residual win value for each individual across all possible categories. While an individual’s true category remains uncertain, observing the sampled opponent’s category allows us to iteratively refine these expected residuals, enabling the inference of the most suitable category.

\begin{figure}[t]
    \centering
    \includegraphics[width=0.25\textwidth]{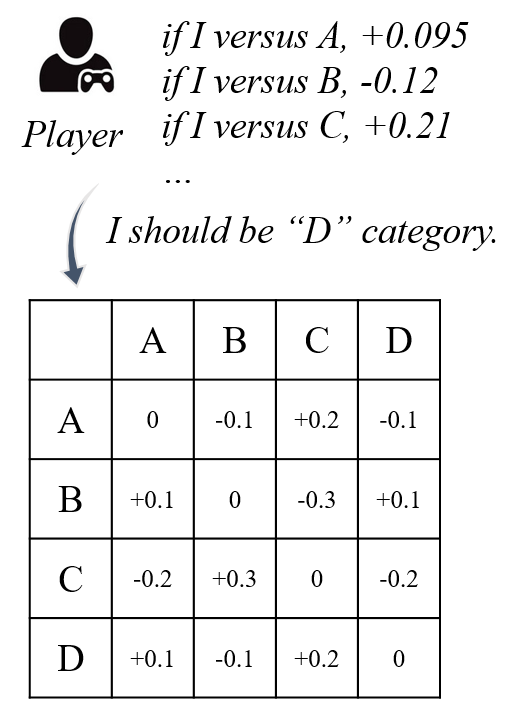}
    \caption{Illustration of the EM algorithm. Residual win values for all categories are learned iteratively, enabling the best-fitting category for each individual to be identified and refined as the classification label.}
    \label{figure:idea}
\end{figure}

This two-step process—learning individual residual values and refining category assignments—can be interpreted as an \textit{expectation-maximization (EM) algorithm} \citep{em_algorithm}. Based on the inferred best category, we refine the categorical distribution for future matches, as illustrated in Figure~\ref{figure:idea}.

To ensure stability in win value prediction, the category with the highest probability is used during inference, rather than sampled categories.

\subsection{Algorithm}

The proposed algorithm follows these steps:
\begin{enumerate}
    \item \textbf{Elo Rating Update}: Update player ratings using the standard Elo formula.
    \item \textbf{Counter Table Update}: Sample counter categories for both players from their categorical distributions and update the $M \times M$ counter table using tabular regression.
    \item \textbf{Expected Residual Update}: Refine the expected residual win values for each individual based on the sampled opponent's category.
    \item \textbf{Category Refinement}: Identify the best-fitting categories by minimizing the discrepancy between the counter table and expected residuals. Update categorical distributions accordingly.
\end{enumerate}

The complete algorithm is presented in Algorithm~\ref{alg:EloRCC}, leveraging softmax for modeling categorical distributions.

\begin{algorithm}
\caption{Online Update Algorithm for Elo Residual Counter Category}
\label{alg:EloRCC}
\begin{algorithmic}[1]
\Require Initial ratings $\mathcal{R}$, category distributions $\mathcal{C}$, counter table $\mathbf{T}$, expected residual table $\mathbf{E}$, learning rates $\eta_R$, $\eta_T$, $\eta_C$, number of categories $M$.
\Ensure Updated ratings $\mathcal{R}$ and category distributions $\mathcal{C}$.

\For{each match $(i, j)$ with result $O_i$ (1 for win, 0 for loss, 0.5 for tie)}
    \LeftComment{Step 1: Elo Rating Update}
    \State $P_i \gets \frac{1}{1 + 10^{(R_j - R_i)/400}}$
    \State $R_i \gets R_i + \eta_R (O_i - P_i)$
    \State $R_j \gets R_j + \eta_R ((1 - O_i) - (1 - P_i))$

    \LeftComment{Step 2: Counter Table Update}
    \State Sample $c_i \sim \mathcal{C}_i$, $c_j \sim \mathcal{C}_j$
    \State $W_{res} \gets O_i - P_i$
    \State $\mathbf{T}[c_i, c_j] \gets \mathbf{T}[c_i, c_j] + \eta_T (W_{res} - \mathbf{T}[c_i, c_j])$
    \State $\mathbf{T}[c_j, c_i] \gets -\mathbf{T}[c_i, c_j]$

    \LeftComment{Step 3: Update Expected Residuals}
    \State $\mathbf{E}_i[c_j] \gets \mathbf{E}_i[c_j] + \eta_T (W_{res} - \mathbf{E}_i[c_j])$
    \State $\mathbf{E}_j[c_i] \gets \mathbf{E}_j[c_i] + \eta_T (-W_{res} - \mathbf{E}_j[c_i])$

    \LeftComment{Step 4: Category Refinement}
    \State $D_i[c] \gets \sum_{c'} |\mathbf{T}[c, c'] - \mathbf{E}_i[c']|$
    \State $D_j[c] \gets \sum_{c'} |\mathbf{T}[c, c'] - \mathbf{E}_j[c']|$
    \State $c_i^* \gets \arg\min_c D_i[c]$
    \State $c_j^* \gets \arg\min_c D_j[c]$
    \LeftComment{$\delta_{c, c_j^*}$ is a one-hot encoded vector:}
    \State $\delta_{c, c_j^*} \gets 
    \begin{cases} 
    1, & \text{if } c = c_j^*, \\ 
    0, & \text{otherwise}.
    \end{cases}$
    \State $\mathcal{C}_i[c] \gets \mathcal{C}_i[c] + \eta_C (\delta_{c, c_i^*} - \mathcal{C}_i[c])$
    \State $\mathcal{C}_j[c] \gets \mathcal{C}_j[c] + \eta_C (\delta_{c, c_j^*} - \mathcal{C}_j[c])$
\EndFor
\end{algorithmic}
\end{algorithm}

This algorithm provides an efficient and interpretable framework for the online learning of ratings and counter categories. Experimental results confirm its ability to achieve comparable or superior performance to NCT in scenarios without complex team combinations.

For a concrete example of how to implement this algorithm, please refer to our Python implementation \footnote{\url{https://github.com/DSobscure/cgi_drl_platform/blob/elo_rcc/platform/cgi_drl/measure_model/elo_rcc/trainer.py}}.

Regarding the space complexity of maintaining this algorithm for $N$ individuals with $M$ counter categories, it consists of the rating table ($\mathcal{O}(N)$), the counter table ($\mathcal{O}(M^2)$), the expected residual win values ($\mathcal{O}(NM)$), and the tabular categorical logits ($\mathcal{O}(NM)$). These components make the method computationally feasible for online applications, particularly when $M$ is not excessively large.

\section{Experiment}
To validate the effectiveness of our proposed algorithm, we adopt the evaluation process from \citet{game_balance_analysis}, utilizing their publicly available datasets and evaluation codes. This ensures a direct comparison with the Neural Rating Table (NRT) and Neural Counter Table (NCT), which serve as benchmarks for our method.

Although these datasets are originally designed for analyzing the strength relationships between card decks or civilization choices to study game balance, they are well-suited for assessing rating systems. Unlike human player ratings, which may involve subjective factors, these datasets provide clear and objective strength relationships with well-documented evidence, allowing for rigorous evaluation of strength prediction accuracy.

\subsection{Datasets and Evaluation Method}

The datasets used in our experiments are collected from two-player zero-sum player-versus-player (PvP) games, including both synthetic toy games and real-world popular online games. Below, we describe each dataset in detail:

\subsubsection{Rock-Paper-Scissors}

The classical Rock-Paper-Scissors dataset consists of 3 individuals (rock, paper, and scissors), with win outcomes represented as $1$, $0$, or $0.5$ for win, lose, or tie, respectively. Matches are generated by uniformly sampling strategies for both players, resulting in a dataset of 100,000 matches.

\subsubsection{Advanced Combination Game}

This synthetic game features an element pool of size 20. Each player selects three unique elements from the pool to form a team. The score $s_c$ of a team $c$ is the sum of its elements. The win probability between two teams is determined by the distribution:$P(c_1>c_2)=\frac{(s_{c_1})^2}{(s_{c_1})^2+(s_{c_2})^2}$.
There are $C^{20}_3 = 1,140$ unique teams in this game. To introduce strategic elements beyond numerical summation, each team is assigned a category $T = s_c \mod 3$, representing Rock ($0$), Paper ($1$), or Scissors ($2$). Teams categorized as Rock, Paper, or Scissors receive a +60 score bonus against their respective weaker category during win-lose sampling. This modification introduces strategy dynamics while preserving numerical selection as a fundamental aspect. The dataset contains 100,000 matches, uniformly sampled across possible team combinations.

\subsubsection{Age of Empires II (AoE2)}

Age of Empires II is a classic and widely popular real-time strategy (RTS) game where players control civilizations to gather resources, build armies, and engage in tactical battles \citep{aoe2,age_of_empires_ii,age_of_empires_ii_steam}. The dataset for this game is derived from match records of 1-on-1 ranked matches, collected and provided by the aoestats website\footnote{\url{https://aoestats.io}}. These records are from patch 99311 and include games played on random maps across all Elo brackets. The dataset features 45 civilizations as testing individuals, with a total of 1,261,288 matches.

\subsubsection{Hearthstone}

Hearthstone is a widely popular collectible card game (CCG) where players build decks and compete in turn-based matches, leveraging strategic card synergies and counterplays \citep{hearthstone}. The dataset for this game focuses on named decks rather than all possible deck configurations. These named decks are collected from the HSReplay platform\footnote{\url{https://hsreplay.net/}}. Specifically, the dataset includes 91 predefined decks as testing individuals, with match records sourced from standard-ranked games at the Gold level in January 2024. In total, the dataset comprises 10,154,929 matches.

\subsubsection{Evaluation Method}

We follow the evaluation protocol outlined in Section 4.2 of the work by \citet{game_balance_analysis}. Pairwise win rates between individuals are computed, and thresholds are used to categorize the strength relations as stronger, weaker, or equal. Specifically:
\begin{itemize}
    \item A win rate within the range $[0.499, 0.501]$ indicates equal strength.
    \item A win rate outside this range classifies one individual as stronger or weaker than the other, depending on the direction.
\end{itemize}

The ground truth for strength relations is derived from these thresholds. The accuracy of a rating method is determined by its ability to predict the same strength relations as the ground truth. All models are trained for 100 epochs (each epoch utilizes the full training set once) using 5-fold cross-validation. The datasets include predefined static splits for cross-validation to ensure consistency.

\subsection{Strength Relation Accuracy}
\begin{table*}
\centering
\caption{Accuracies (\%) for different rating systems in training (top) and testing (bottom) sets using $M=81$. Results are averaged over 5-fold cross-validation, with $\pm$ one standard deviation. $\dagger$ Results for NRT, NCT, and Elo are reproduced from \citet{game_balance_analysis}. Elo-RCC achieves comparable performance to NCT in games without complex synergy, such as Rock-Paper-Scissors and Age of Empires II, while outperforming other online update methods.}

\label{table:precision_test1}
\begin{tabular}{lll|llll}
\toprule
\quad & \textbf{NRT$^\dagger$} & \textbf{NCT$^\dagger$} M=81& \textbf{Elo$^\dagger$} & \textbf{mElo2$^\dagger$} & \textbf{Elo} K=0.1 & \textbf{Elo-RCC} M=81 \\
\midrule
Rock-Paper-Scissors & 51.3 $\pm$ 10.0 & \textbf{100} $\pm$ 0 & 73.3 $\pm$ 10.1 & \textbf{100} $\pm$ 0 & 60.0 $\pm$ 9.9 & \textbf{100} $\pm$ 0\\
\midrule
Advanced Combination & 57.9 $\pm$ 0 & \textbf{79.4} $\pm$  0.8 & 57.1 $\pm$ 0.2 & 52.7 $\pm$ 2.8 & 57.7 $\pm$ 0.1 & 68.0 $\pm$ 0.3\\
\midrule
Age of Empires II & 68.7 $\pm$ 0.8 & \textbf{97.7} $\pm$ 1.0 & 53.1 $\pm$ 4.1 & 51.2 $\pm$ 3.7 & 68.5 $\pm$ 1.0 & \textbf{94.3} $\pm$ 1.4\\
\midrule
Hearthstone & 83.4 $\pm$ 4.6 & \textbf{97.4} $\pm$ 0.3 & 74.8 $\pm$ 2.6 & 61.4 $\pm$ 5.6 & 81.6 $\pm$ 0.5 & \textbf{95.9} $\pm$ 0.9\\
\bottomrule
\toprule
Rock-Paper-Scissors & 51.1 $\pm$ 9.9 & \textbf{100} $\pm$  0 & 73.6 $\pm$ 9.9 & \textbf{100} $\pm$ 0 & 60.2 $\pm$ 10.2 & \textbf{100} $\pm$ 0\\
\midrule
Advanced Combination & 56.5 $\pm$ 0.3 & \textbf{79.7} $\pm$ 0.8 & 56.1 $\pm$ 0.2 & 51.3 $\pm$ 0.6 & 56.4 $\pm$ 0.3 & 65.3 $\pm$ 0.6\\
\midrule
Age of Empires II & 64.5 $\pm$ 1.3 & \textbf{75.4} $\pm$ 0.9 & 52.4 $\pm$ 4.9 & 51.0 $\pm$ 3.1 & 64.7 $\pm$ 1.7& \textbf{75.0} $\pm$ 1.2\\
\midrule
Hearthstone & 81.2 $\pm$ 0.5 & \textbf{94.8} $\pm$ 0.5 & 74.9 $\pm$ 2.6 & 61.1 $\pm$ 5.8 & 81.3 $\pm$ 0.7 & \textbf{94.4} $\pm$ 0.8\\
\bottomrule
\end{tabular}
\end{table*}

\begin{table*}[!htb]
\centering
\caption{Accuracies (\%) for smaller counter category sizes ($M=3, 9, 27$) in training (top) and testing (bottom) sets. Results are averaged over 5-fold cross-validation, with $\pm$ one standard deviation. $\dagger$ Results for NCT are reproduced from \citet{game_balance_analysis}. Elo-RCC demonstrates slightly higher accuracy than NCT for $M=3$ and $M=9$, showcasing its robustness in smaller discretization scenarios.}

\label{table:precision_test2}
\begin{tabular}{l|ll|ll|ll}
\toprule
\quad & M=3 \textbf{NCT$^\dagger$} & \textbf{Elo-RCC} & M=9 \textbf{NCT$^\dagger$} & \textbf{Elo-RCC} & M=27 \textbf{NCT$^\dagger$} & \textbf{Elo-RCC} \\
\midrule
Rock-Paper-Scissors & \textbf{100} $\pm$ 0 & \textbf{100} $\pm$ 0 & \textbf{100} $\pm$ 0 & \textbf{100} $\pm$ 0 & \textbf{100} $\pm$ 0 & \textbf{100} $\pm$ 0\\
\midrule
Advanced Combination & \textbf{57.9} $\pm$ 0.0 & \textbf{58.1} $\pm$ 0.8 & \textbf{79.4} $\pm$ 0.6 & 57.7 $\pm$ 0.1 & \textbf{79.8} $\pm$ 0.2 & 57.7 $\pm$ 0.1\\
\midrule
Age of Empires II & \textbf{68.7} $\pm$ 0.8 & \textbf{71.0} $\pm$ 0.5 & \textbf{73.1} $\pm$ 3.9 & \textbf{74.6} $\pm$ 1.2 & \textbf{83.8} $\pm$ 1.4 & \textbf{83.6} $\pm$ 1.4\\
\midrule
Hearthstone & \textbf{81.3} $\pm$ 0.2 & \textbf{81.6} $\pm$ 1.2 & \textbf{85.4} $\pm$ 1.2 & \textbf{85.9} $\pm$ 0.8 & \textbf{91.7} $\pm$ 0.6 & \textbf{91.2} $\pm$ 1.1\\
\bottomrule
\toprule
Rock-Paper-Scissors & \textbf{100} $\pm$ 0 & \textbf{100} $\pm$ 0 & \textbf{100} $\pm$ 0 & \textbf{100} $\pm$ 0 & \textbf{100} $\pm$ 0 & \textbf{100} $\pm$ 0\\
\midrule
Advanced Combination & \textbf{56.5} $\pm$ 0.3 & \textbf{56.4} $\pm$ 0.3 & \textbf{79.7} $\pm$ 0.4 & 56.4 $\pm$ 0.3 & \textbf{80.1} $\pm$ 0.4 & 56.4 $\pm$ 0.3\\
\midrule
Age of Empires II & \textbf{64.5} $\pm$ 1.3 & \textbf{66.5} $\pm$ 2.1 & \textbf{67.7} $\pm$ 2.5 & \textbf{68.7} $\pm$ 1.8 & \textbf{72.5} $\pm$ 1.1 & \textbf{71.3} $\pm$ 0.5\\
\midrule
Hearthstone & \textbf{81.3} $\pm$ 0.5 & \textbf{81.5} $\pm$ 1.5 & \textbf{85.2} $\pm$ 0.9 & \textbf{85.7} $\pm$ 1.3 & \textbf{91.3} $\pm$ 0.8 & \textbf{90.9} $\pm$ 1.1\\
\bottomrule
\end{tabular}
\end{table*}

In this section, we evaluate the performance of our proposed Elo-RCC algorithm against several baseline methods. To assess the impact of different counter category sizes, we test $M = 3, 9, 27, 81$, aligning with the configurations used in NCT. The methods compared include:

\begin{itemize}
    \item \textbf{NRT}: The Neural Rating Table, a neural network-based extension of the Bradley-Terry model. While it effectively models synergies between players or teams, it cannot address win value intransitivity.
    
    \item \textbf{NCT}: The Neural Counter Table, building on NRT, incorporates an $M \times M$ counter table to handle intransitive relationships using neural networks. Results are evaluated for $M = 3, 9, 27,$ and $81$.
    
    \item \textbf{Elo}: The standard Elo rating system, initialized with a rating of 1000 and an update constant $K = 16$. This is a widely used baseline for measuring player strength in games.
    
    \item \textbf{mElo2}: A multi-dimensional extension of the Elo system \citep{m_elo}, which models intransitive relationships using a vector representation. It is initialized with a rating of 1000 and an update constant $K = 16$.
    
    \item \textbf{Elo with small $K$}: A variant of the Elo system with a smaller update constant $K = 0.1$, designed to improve accuracy in scenarios with minor strength differences, such as in Age of Empires II matches.
    
    \item \textbf{Elo-RCC}: Our proposed method, initialized with a rating of 1000 and incorporating additional parameters for counter category learning. The update rates are set as follows: $\eta_R = 0.1$ (analogous to $K$ in Elo), $\eta_T = 0.00025$ for updating the counter table, and $\eta_C = 0.01$ for refining category distributions. We test Elo-RCC with the same category sizes as NCT $( M = 3, 9, 27,$ and $81)$.
\end{itemize}

Table~\ref{table:precision_test1} presents the strength relation accuracy across different rating systems for $M = 81$ (NCT and Elo-RCC). As expected, NCT achieves the highest accuracy in most cases due to its neural network-based design. However, Elo-RCC closely matches NCT's performance in games without complex synergies, such as Rock-Paper-Scissors and Age of Empires II, demonstrating its effectiveness in handling intransitive relationships without relying on neural networks.

When evaluating smaller counter table sizes ($M = 3, 9, 27$) in Table~\ref{table:precision_test2}, Elo-RCC slightly outperforms NCT for $M = 3$ and $M = 9$. This is likely because the discretization process in NCT struggles when the codebook size is small, whereas Elo-RCC directly adjusts the counter table in real time, providing better adaptability for smaller $M$ values. For larger $M$, such as $M = 27$, Elo-RCC maintains comparable performance to NCT, validating its scalability and robustness.

In summary, our experimental results show that Elo-RCC provides near-equivalent accuracy to NCT for most scenarios, while offering the advantages of online updates and computational simplicity. This makes Elo-RCC a practical alternative for real-time rating systems in games and multi-agent environments.

\section{Conclusion and Future Works}

In this paper, we proposed Elo Residual Counter Category (Elo-RCC), a novel rating system that extends the Neural Counter Table (NCT) to support online updates with tabular approximations. The core innovation lies in modeling counter relationships through categorical distributions and leveraging an EM algorithm to iteratively refine pseudo-category labels. Experimental results demonstrate the effectiveness of our algorithm, particularly in popular online games, where it achieves comparable performance to NCT while offering a lightweight, real-time solution suitable for game balance analysis.

The applications of rating systems extend far beyond player ranking and matchmaking in games. They can be broadly applied to any competitive scenario, including image or video preference ranking, sports analysis, movie recommendation, peer grading, elections, and even the evaluation of large language models \citep{beauty_score, nd_ability, chatbot_arena}. By providing a simpler and more accessible implementation of NCT, Elo-RCC has the potential to facilitate a wider range of applications in these and other domains.

Future work could explore further enhancements to Elo-RCC, such as adapting it for more complex team compositions, incorporating dynamic counter category sizes, or integrating it with neural network models for hybrid approaches. Additionally, investigating its performance in asymmetric or non-zero-sum games and extending its use cases to multi-agent learning frameworks could unlock new avenues for research and practical applications.

\bibliography{main}

\begin{thebibliography}{27}
\providecommand{\natexlab}[1]{#1}
\providecommand{\url}[1]{\texttt{#1}}
\expandafter\ifx\csname urlstyle\endcsname\relax
  \providecommand{\doi}[1]{doi: #1}\else
  \providecommand{\doi}{doi: \begingroup \urlstyle{rm}\Url}\fi

\bibitem[Balduzzi et~al.(2018)Balduzzi, Tuyls, P{\'{e}}rolat, and Graepel]{m_elo}
David Balduzzi, Karl Tuyls, Julien P{\'{e}}rolat, and Thore Graepel.
\newblock Re-evaluating evaluation.
\newblock In \emph{Advances in Neural Information Processing Systems ({NeurIPS})}, 2018.

\bibitem[{Blizzard Entertainment}(2014)]{hearthstone}
{Blizzard Entertainment}.
\newblock Hearthstone, 2014.
\newblock URL \url{https://hearthstone.blizzard.com/}.

\bibitem[Bradley \& Terry(1952)Bradley and Terry]{bradley_terry}
Ralph~Allan Bradley and Milton~E. Terry.
\newblock Rank analysis of incomplete block designs: I. the method of paired comparisons.
\newblock \emph{Biometrika}, 1952.

\bibitem[Bromley et~al.(1993)Bromley, Bentz, Bottou, Guyon, LeCun, Moore, S{\"{a}}ckinger, and Shah]{siamese_network}
Jane Bromley, James~W. Bentz, L{\'{e}}on Bottou, Isabelle Guyon, Yann LeCun, Cliff Moore, Eduard S{\"{a}}ckinger, and Roopak Shah.
\newblock Signature verification using {A} "{S}iamese" time delay neural network.
\newblock \emph{International Journal of Pattern Recognition and Artificial Intelligence ({IJPRAI})}, 1993.

\bibitem[Chen \& Joachims(2016)Chen and Joachims]{nd_ability}
Shuo Chen and Thorsten Joachims.
\newblock Modeling intransitivity in matchup and comparison data.
\newblock In \emph{International Conference on Web Search and Data Mining ({WSDM})}, 2016.

\bibitem[Coulom(2008)]{whr}
R{\'{e}}mi Coulom.
\newblock Whole-history rating: {A} bayesian rating system for players of time-varying strength.
\newblock In \emph{International conference on computers and games ({CG})}, 2008.

\bibitem[Dempster et~al.(1977)Dempster, Laird, and Rubin]{em_algorithm}
Arthur~P Dempster, Nan~M Laird, and Donald~B Rubin.
\newblock Maximum likelihood from incomplete data via the em algorithm.
\newblock \emph{Journal of the royal statistical society: series B (methodological)}, 1977.

\bibitem[Ebtekar \& Liu(2021)Ebtekar and Liu]{elo_mmr}
Aram Ebtekar and Paul Liu.
\newblock Elo-mmr: {A} rating system for massive multiplayer competitions.
\newblock In \emph{The Web Conference ({WWW})}, 2021.

\bibitem[Elo(1966)]{elo}
Arpad~E. Elo.
\newblock \emph{The USCF Rating System: Its Development, Theory, and Applications}.
\newblock United States Chess Federation, 1966.

\bibitem[{Ensemble Studios}(1999)]{aoe2}
{Ensemble Studios}.
\newblock Age of empires ii, 1999.
\newblock URL \url{https://www.ageofempires.com/}.

\bibitem[{Forgotten Empires} et~al.(2019{\natexlab{a}}){Forgotten Empires}, {Tantalus Media}, and {Wicked Witch}]{age_of_empires_ii}
{Forgotten Empires}, {Tantalus Media}, and {Wicked Witch}.
\newblock Age of empires ii: Definitive edition, 2019{\natexlab{a}}.
\newblock URL \url{https://www.ageofempires.com/games/aoeiide/}.

\bibitem[{Forgotten Empires} et~al.(2019{\natexlab{b}}){Forgotten Empires}, {Tantalus Media}, and {Wicked Witch}]{age_of_empires_ii_steam}
{Forgotten Empires}, {Tantalus Media}, and {Wicked Witch}.
\newblock Age of empires ii: Definitive edition (steam), 2019{\natexlab{b}}.
\newblock URL \url{https://store.steampowered.com/app/813780/Age_of_Empires_II_Definitive_Edition/}.
\newblock Available on Steam.

\bibitem[Fujii(2024)]{neural_bt_model}
Satoru Fujii.
\newblock Neural bradley-terry rating: Quantifying properties from comparisons.
\newblock In \emph{International Conference on Agents and Artificial Intelligence {(ICAART)}}, 2024.

\bibitem[Glickman(1999)]{glicko}
Mark~E Glickman.
\newblock Parameter estimation in large dynamic paired comparison experiments.
\newblock \emph{Journal of the Royal Statistical Society Series C: Applied Statistics}, 1999.

\bibitem[Herbrich et~al.(2006)Herbrich, Minka, and Graepel]{true_skill}
Ralf Herbrich, Tom Minka, and Thore Graepel.
\newblock True{S}kill™: A {B}ayesian skill rating system.
\newblock In \emph{Advances in Neural Information Processing Systems ({NIPS})}, 2006.

\bibitem[Hunter(2004)]{generalized_bt}
David~R Hunter.
\newblock Mm algorithms for generalized bradley-terry models.
\newblock \emph{The annals of statistics}, 2004.

\bibitem[Li et~al.(2021)Li, Ma, and Hu]{beauty_score}
Shiyu Li, Hao Ma, and Xiangyu Hu.
\newblock Neural image beauty predictor based on bradley-terry model.
\newblock \emph{CoRR}, abs/2111.10127, 2021.

\bibitem[Lin et~al.(2021)Lin, Chiu, and Wu]{playstyle_distance}
Chiu{-}Chou Lin, Wei{-}Chen Chiu, and I{-}Chen Wu.
\newblock An unsupervised video game playstyle metric via state discretization.
\newblock In \emph{Conference on Uncertainty in Artificial Intelligence ({UAI})}, 2021.

\bibitem[Lin et~al.(2024{\natexlab{a}})Lin, Chiu, and Wu]{playstyle_similarity}
Chiu-Chou Lin, Wei-Chen Chiu, and I-Chen Wu.
\newblock Perceptual similarity for measuring decision-making style and policy diversity in games.
\newblock \emph{Transactions on Machine Learning Research}, 2024{\natexlab{a}}.

\bibitem[Lin et~al.(2024{\natexlab{b}})Lin, Shih, Kuo, Chen, Chen, Chiu, and Wu]{game_balance_analysis}
Chiu-Chou Lin, Yu-Wei Shih, Kuei-Ting Kuo, Yu-Cheng Chen, Chien-Hua Chen, Wei-Chen Chiu, and I-Chen Wu.
\newblock Identifying and clustering counter relationships of team compositions in pvp games for efficient balance analysis.
\newblock \emph{Transactions on Machine Learning Research}, 2024{\natexlab{b}}.

\bibitem[Pendurkar et~al.(2023)Pendurkar, Chow, Jie, and Sharon]{nash_entropy_balancing}
Sumedh Pendurkar, Chris Chow, Luo Jie, and Guni Sharon.
\newblock Bilevel entropy based mechanism design for balancing meta in video games.
\newblock In \emph{International Conference on Autonomous Agents and Multiagent Systems ({AAMAS})}, 2023.

\bibitem[Pramono et~al.(2018)Pramono, Renalda, and Warnars]{mmr}
Muhammad~Farrel Pramono, Kevin Renalda, and Harco Leslie Hendric~Spits Warnars.
\newblock Matchmaking problems in moba games.
\newblock \emph{Indonesian Journal of Electrical Engineering and Computer Science}, 2018.

\bibitem[Schell(2008)]{art_of_game_design}
Jesse Schell.
\newblock \emph{The Art of Game Design: A book of lenses}.
\newblock CRC press, 2008.

\bibitem[Vadori \& Savani(2024)Vadori and Savani]{hyperbolic_elo}
Nelson Vadori and Rahul Savani.
\newblock Ordinal potential-based player rating.
\newblock In \emph{International Conference on Artificial Intelligence and Statistics ({AISTATS})}, 2024.

\bibitem[van~den Oord et~al.(2017)van~den Oord, Vinyals, and Kavukcuoglu]{vq_vae}
A{\"{a}}ron van~den Oord, Oriol Vinyals, and Koray Kavukcuoglu.
\newblock Neural discrete representation learning.
\newblock In \emph{Advances in Neural Information Processing Systems ({NeurIPS})}, 2017.

\bibitem[Vinyals et~al.(2019)Vinyals, Babuschkin, Czarnecki, Mathieu, Dudzik, Chung, Choi, Powell, Ewalds, Georgiev, Oh, Horgan, Kroiss, Danihelka, Huang, Sifre, Cai, Agapiou, Jaderberg, Vezhnevets, Leblond, Pohlen, Dalibard, Budden, Sulsky, Molloy, Paine, Gulcehre, Wang, Pfaff, Wu, Ring, Yogatama, W{\"u}nsch, McKinney, Smith, Schaul, Lillicrap, Kavukcuoglu, Hassabis, Apps, and Silver]{alpha_star}
Oriol Vinyals, Igor Babuschkin, Wojciech~M. Czarnecki, Micha{\"e}l Mathieu, Andrew Dudzik, Junyoung Chung, David~H. Choi, Richard Powell, Timo Ewalds, Petko Georgiev, Junhyuk Oh, Dan Horgan, Manuel Kroiss, Ivo Danihelka, Aja Huang, L.~Sifre, Trevor Cai, John~P. Agapiou, Max Jaderberg, Alexander~Sasha Vezhnevets, R{\'e}mi Leblond, Tobias Pohlen, Valentin Dalibard, David Budden, Yury Sulsky, James Molloy, Tom~Le Paine, Caglar Gulcehre, Ziyun Wang, Tobias Pfaff, Yuhuai Wu, Roman Ring, Dani Yogatama, Dario W{\"u}nsch, Katrina McKinney, Oliver Smith, Tom Schaul, Timothy~P. Lillicrap, Koray Kavukcuoglu, Demis Hassabis, Chris Apps, and David Silver.
\newblock Grandmaster level in starcraft ii using multi-agent reinforcement learning.
\newblock \emph{Nature}, 2019.

\bibitem[Zheng et~al.(2023)Zheng, Chiang, Sheng, Zhuang, Wu, Zhuang, Lin, Li, Li, Xing, Zhang, Gonzalez, and Stoica]{chatbot_arena}
Lianmin Zheng, Wei{-}Lin Chiang, Ying Sheng, Siyuan Zhuang, Zhanghao Wu, Yonghao Zhuang, Zi~Lin, Zhuohan Li, Dacheng Li, Eric~P. Xing, Hao Zhang, Joseph~E. Gonzalez, and Ion Stoica.
\newblock Judging llm-as-a-judge with mt-bench and chatbot arena.
\newblock In \emph{Advances in Neural Information Processing Systems ({NeurIPS})}, 2023.

\end{thebibliography}

\end{document}